\documentclass[preprint,5p]{elsarticle}

\usepackage{graphicx} 
\graphicspath{ {./images/} }
\usepackage{amssymb} 
\usepackage{amsthm} 
\usepackage{float} 
\usepackage{lineno} 
\usepackage{soul,color}
\usepackage{subcaption}
\usepackage{multirow}
\usepackage{amsmath}
\usepackage{hyperref}
\usepackage{lineno}
\usepackage{makecell}

\journal{Engineering Structures}

\begin{document}

\begin{frontmatter}

\title{Virtual Axle Detector based on Analysis of Bridge Acceleration Measurements by Fully Convolutional Network}



\author[label1]{Steven Robert Lorenzen\corref{cor1}}
\author[label1]{Henrik Riedel}
\author[label1]{Maximilian Michael Rupp}
\author[label1]{Leon Schmeiser}
\author[label1]{Hagen Berthold}
\author[label2]{Andrei Firus}
\author[label1]{Jens Schneider}

\cortext[cor1]{
    Corresponding author. \textit{E-mail address:} lorenzen@ismd.tu-darmstadt.de (S. R. Lorenzen).
    }

\address[label1]{Institute for Structural Mechanics and Design, TU Darmstadt, Germany}
\address[label2]{iSEA Tec GmbH, Germany}

\begin{abstract}
In the practical application of the Bridge Weigh-In-Motion (BWIM) methods, the position of the wheels or axles during the passage of a vehicle is in most cases a prerequisite. To avoid the use of conventional axle detectors and bridge type specific methods, we propose a novel method for axle detection through the placement of accelerometers at any point of a bridge. In order to develop a model that is as simple and comprehensible as possible, the axle detection task is implemented as a binary classification problem instead of a regression problem. The model is implemented as a Fully Convolutional Network to process signals in the form of Continuous Wavelet Transforms. This allows passages of any length to be processed in a single step with maximum efficiency while utilising multiple scales in a single evaluation. This enables our method to use acceleration signals at any location of the bridge structure serving as Virtual Axle Detectors (VADs) without being limited to specific structural types of bridges. To test the proposed method, we analysed 3787 train passages recorded on a steel trough railway bridge of a long-distance traffic line. Our results on the measurement data show that our model detects 95\% of the axes, thus, 128,599 of 134,800 previously unseen axles were correctly detected. In total, 90\% of the axles can be detected with a maximum spatial error of 20~cm, with a maximum velocity of $v_{\mathrm{max}}=56,3~\mathrm{m/s}$. The analysis shows that our developed model can use accelerometers as VADs even under real operating conditions. 
\end{abstract}

\end{frontmatter}



\newpage
\section{Introduction}
\label{S:1}

All over the world, ageing bridge infrastructure faces the challenge of increasing traffic loads. For example, in the United States, there are more than 617,000 bridges, 42\% of which are at least 50 years old and 7.5\%  of them are considered structurally deficient \cite{ASCE2021}. In Germany, more than 40\% of the 25,710 railway bridges are older than 80 years, while the average lifespan is about 122 years \cite{Geissler2014,Knapp2019_2}. The application of structural health monitoring (SHM) makes it possible to increase the operational availability and safety of these structures. As the knowledge of the actual operational loads is of high importance for the condition assessment of the structures, especially when it comes down to the assessment of fatigue failure and the evaluation of the remaining service life, the determination of the loads is a key aspect in the field of SHM. Since direct measurement of the loads is often technically difficult and usually requires a significant financial effort \cite{Chan2001,Kouroussis2015, Firus2022}, different  methods for load identification based on measured structural responses  have been developed \cite{KazemiAmiri2017,Hwang2009, Lourens2012,Firus2021,Firus2022}. In the case of bridges, these techniques are referred to as Bridge Weigh-In-Motion (BWIM) \cite{Lydon2017d,Wang2019,Yu2016b,He2019}. 

For the majority of BWIM systems, the information on vehicle configuration (axle number and axle spacing) and velocity are prerequisites \cite{He2019}. For this purpose, conventional axle detectors are used \cite{Thater1998,Zakharenko2022,Bernas2018,Kouroussis2015}. However, due to impact loads of the wheels, the axle detectors have a limited durability \cite{Yu2017}. In addition, the installation of the axle detectors always implies road or railway track closures. Especially the latter case requires a considerable bureaucratic, logistic and financial effort. To avoid these issues, modern BWIM systems use axle detection concepts that only use sensors installed under the bridge. These concepts are referred as nothing-on-road (NOR) or free-of-axle-detector (FAD) \cite{OBrien2012,Zhao2020}.

In the FAD technology,  two additional strain sensors at different positions of the bridge are used to identify the vehicle configuration and velocity \cite{Zhao2020}. Since FAD is only suitable for specific types of bridges \cite{He2019}, researchers attempted to identify axle velocity and spacing from global flexural strain or shear strain measurements \cite{He2019, Kalhori2017}. For the proposed method in \cite{Kalhori2017}, the use of shear strains, require the application of the stain gauges at the level of the neutral axis. This is a challenge for complex structures, especially for railway bridges with ballasted tracks, as in such cases, the position of the neutral axis cannot be easily determined.  However, in \cite{He2019}, the proposed method is only  suitable for structures where the structural response is dominated by the quasi static response of the bridge, e.g. where the dynamic amplification is low. Furthermore, in \cite{He2019}, the second time derivative of the strain signals is used. This makes the method sensitive to measurement noise, which leads to the necessity of a suitable noise filter, depending on the specific application. 

In \cite{He2019} the method of virtual axles is proposed, in which a vehicle with many virtual axles is assumed. All axles except the real ones are weightless. The true axles and their weights are then determined by solving a constrained least square problem. As the authors state, the method fails, if there is significant noise in the signals. Since a significant amount of noise is present in field measurements and further practical applications, the method cannot be practically applied without a sophisticated regularisation method. Furthermore, the method uses experimentally determined lines of influence, so it is not applicable to cases with significant dynamic amplification in their structural response. 

To the best of our knowledge, only \citet{Zhu2021} have so far published an accelerometer-based axle detection method. Here, a shallow Convolutional Neural Network (CNN) is used to detect potential axle sequences, which are then transformed with a continuous wavelet transform. Afterwards, the axles are detected in the transformed signals by peak-finding methods. The method of \citet{Zhu2021} requires accelerometers close to the supports. The acceleration signals of these sensors are dominated by the vehicle induced impulses when entering and leaving the bridge, leading to the clear axle recognition in the time domain.

An axle detection method based on acceleration measurements is desirable, as the installation of acceleration sensors is much easier and less laborious, compared to to strain gauges. However, accelerometers are often already installed on the structure for the determination of the modal parameters and they are not necessarily located close to the supports.

Therefore, we propose a method that enables the use of accelerometers attached at any position on a bridge as a VAD. In this way, the same acceleration sensors used for analyzing the global structural behavior (e.g. at midspan or quarter span of beam-like bridges) can be employed also for the axle detection, without having to install additional sensors in the proximity of the supports.
Continuous-Wavelet-Transformations (CWTs) were used in the present work as they are an effective tool for the analysis of acoustic and visual signals in general \citep{DaubechiesCWT}, furthermore  previous work \citep{Chatterjee2006, Kalhori2017, Yu2017, Zhao2020, Zhu2021} have shown that CWTs are an effective tool for axle identification.
The wavelet transformed signals are subsequently analysed using a Fully Convolutional Network (FCN) that is trained in a supervised manner to perform a binary classification task (axle/no axle). This enables the signal processing of any length without having to be divided into time windows. Furthermore, the analysis in this way is not limited to certain mother wavelets or certain scales, as in the previously mentioned works that used the CWT \citep{Chatterjee2006, Kalhori2017, Yu2017, Zhao2020, Zhu2021}. To validate our method, we recorded a data set on a railway bridge with sensors distributed across the free span of the bridge on the main girders. The impulses of the wheel sets are superimposed with the vibrations of the bridge, which does not allow their clear visual identification in the time domain. For many bridges and for common sensor setups for monitoring purposes, similar to the ones used for this study, the method of \citet{Zhu2021} would not be applicable. 

Since we use a supervised learning approach for the VAD, a set of train passages with known axle distances and velocity are required. In the current research, this information is obtained by means of strain measurements at the rail level. For future practical applications, the information could be obtained from vehicles with known axle configuration and the use of a Differential Global Positioning System (DGPS). If such information is not available, a transfer learning approach based on simulated data could also be an option.

The paper is structured as follows: In chapter two, the methods are presented. The first section of the chapter describes the data acquisition in the field experiment and the subsequent data processing. The second section of chapter two contains the model definition. In the last section of the chapter, details are given on the training of the model. Chapter three presents and discusses the results. The paper ends with chapter four, in which the conclusions of the present study are drawn.
\section{Methods}
\label{S:2}
\subsection{Data Acquisition}

\begin{figure}[H]
\centering\includegraphics[trim = 0mm 50mm 0mm 50mm,clip,width=\linewidth]{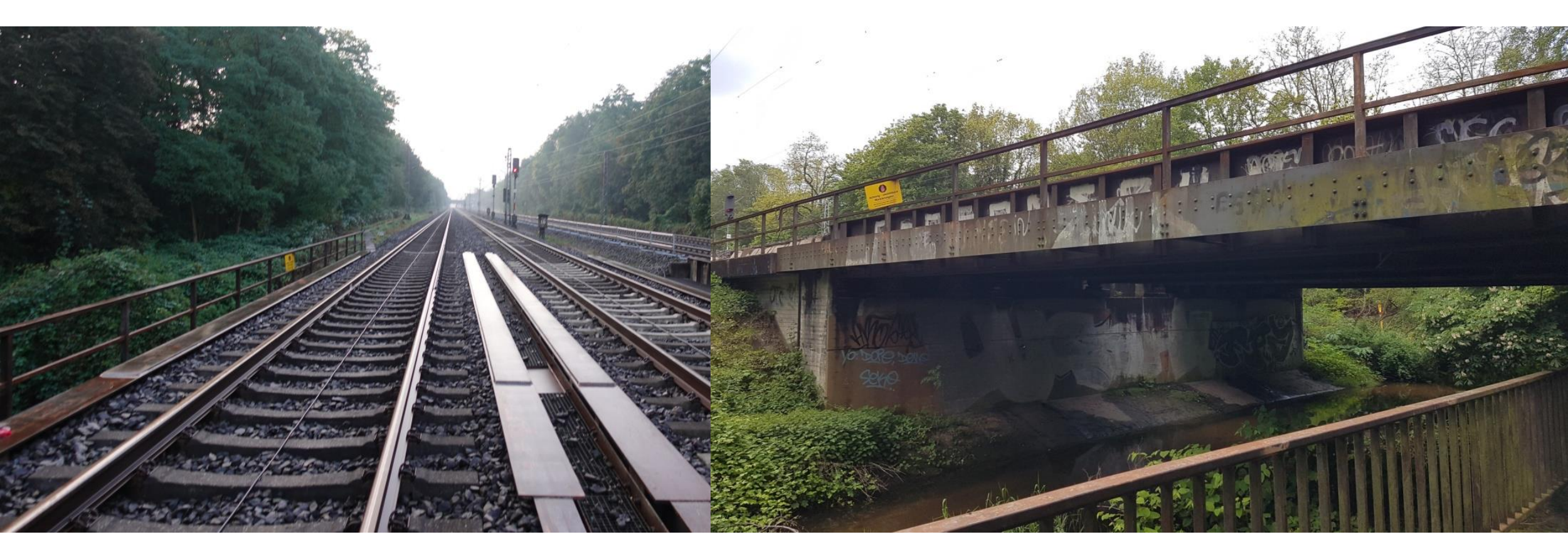}
\caption{Photos of the investigated structure}
\label{F:BridgePhoto}
\end{figure}

\begin{figure*}
\centering\includegraphics[trim = 0mm 40mm 185mm 13mm,clip,width=\textwidth]{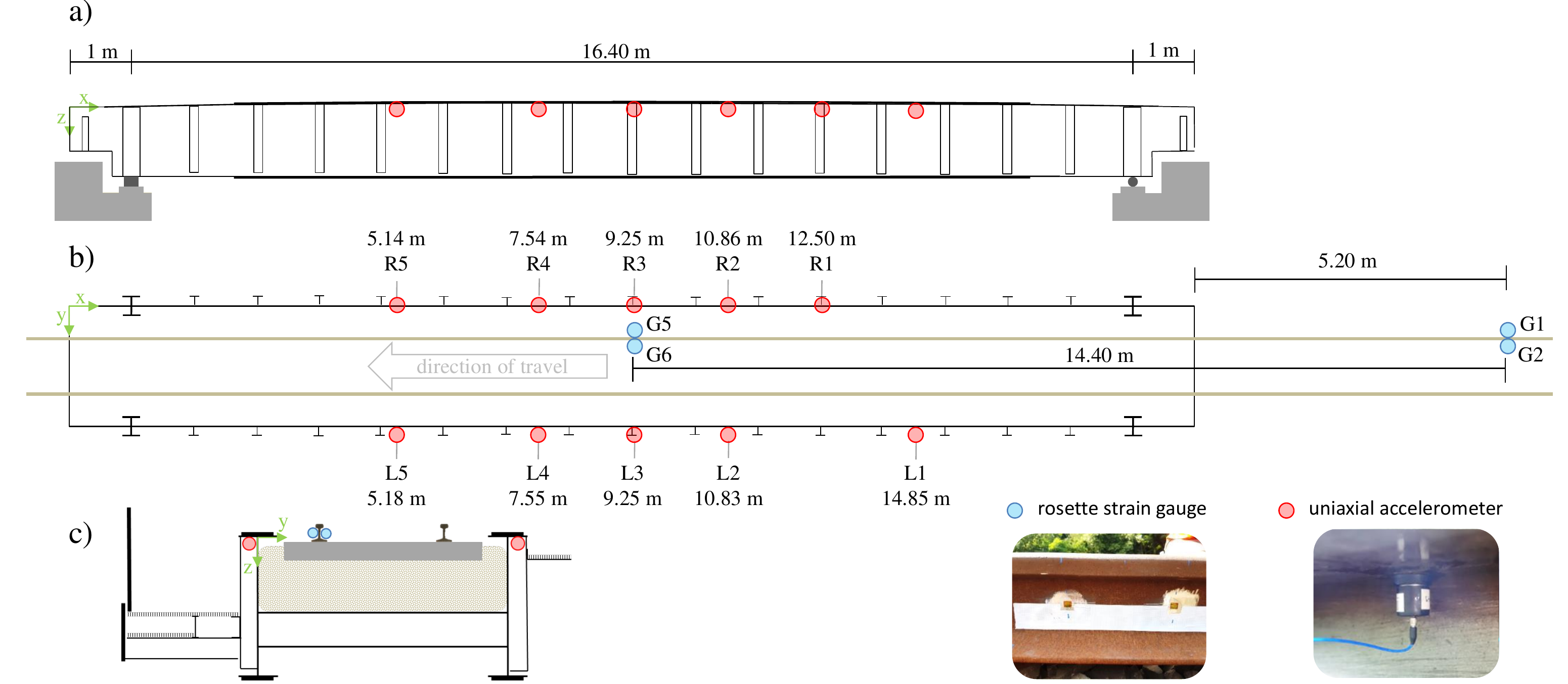}
\caption{Bridge and sensor setup a) side view b) top view with sensor labels, accelerometers $x$-ordinate and strain gauge distances  c) cross section}
\label{F:BridgeSensorsetup}
\end{figure*}

We recorded the measurement data used in the present study on a single-span steel trough railway bridge (fig.~\ref{F:BridgePhoto}) located on a long-distance traffic line in Germany. The bridge has a total length of 18.4~m with a free span of 16.4~m (fig.~\ref{F:BridgeSensorsetup}).

The measurement set-up is shown in fig.~\ref{F:BridgeSensorsetup}. It can be seen that a total of  ten seismic uniaxial accelerometers of the type PCB-39B04 (PCB Synotech) with a sensitivity of 1000~mV/g~(±10\%), a broadband resolution of 0.000003~g$_\mathrm{RMS}$, a measurement range of ±5~g$_\mathrm{pk}$ and a frequency range of 0.06~to~450~Hz~(±5\%) were installed. 

The measurements are triggered via the rising slope of the wheel load measuring point G1 (fig.~\ref{F:BridgeSensorsetup}) from the ring buffer are stored from 10 seconds before the trigger together with the 50 seconds long measurement after the triggering. The recorded signals thus all have a length of 60 seconds. All sensor signals were recorded with a sampling frequency of $f_{\mathrm{s}}=600~\mathrm{Hz}$ using the catmanAP software and the CX22 data recorder connected to an MX1601B universal amplifier and an MX1616B strain gauge amplifier (all products are from HBK).

\begin{figure*}
\centering\includegraphics[trim = 20mm 10mm 20mm 10mm,clip,width=\textwidth]{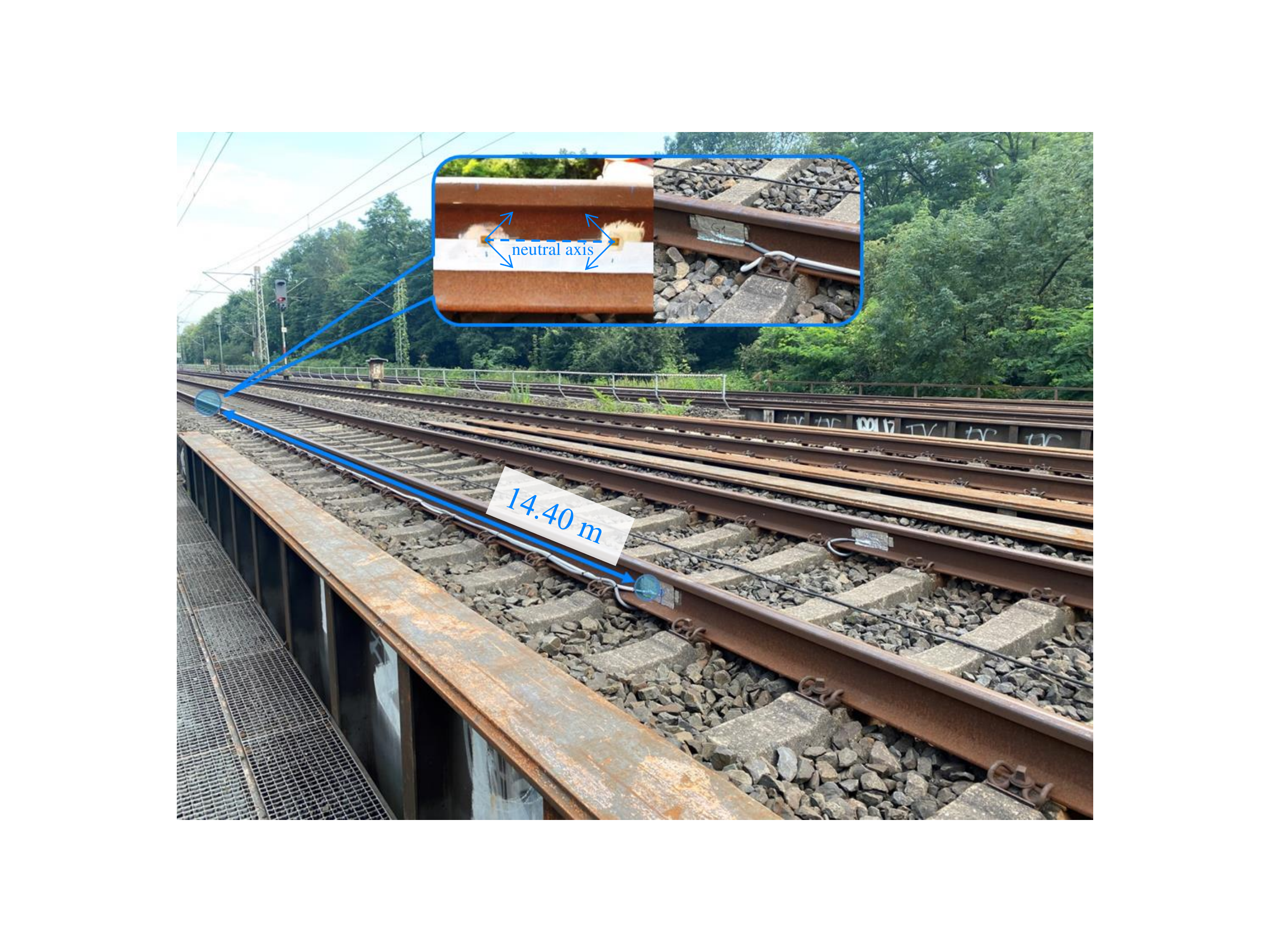}
\caption{Photo with marking of the wheel load measuring points used and their spacing. The detail section shows the position of the rosette strain gauges and the installed weatherproof measuring point.}
\label{F:ALMPhoto}
\end{figure*}

By means of two wheel load measuring points, the average velocity of each axle was determined and from this the actual position of the axles during the passage was deduced. Every measuring point involves installation of at least one pair of rosette strain gauges (HBM 1-CXY41-6/350HE) on the rails. Each pair of strain gauges are placed at the level of the neutral axis of the UIC~60 rail profiles with a distance of 20~cm and allow the recording of bi-axial strains at an angle of 45° with respect to the neutral axis (fig.~\ref{F:ALMPhoto}). Thus shear strains are obtained. The difference of the shear strains allows the determination of the acting wheel loads. Since only the difference of the shear strains is of interest, the strain gauges can be combined into a single signal in a full bridge circuit. For further details, please refer to e.g. \cite{Kouroussis2015}. To compensate for the influence of the lateral wheel loads, a pair of strain gauges was placed on each side of the rail, so that one wheel load measuring point retrieves two signals.

\begin{figure*}
\centering\includegraphics[trim = 0mm 40mm 0mm 20mm,clip,width=\linewidth]{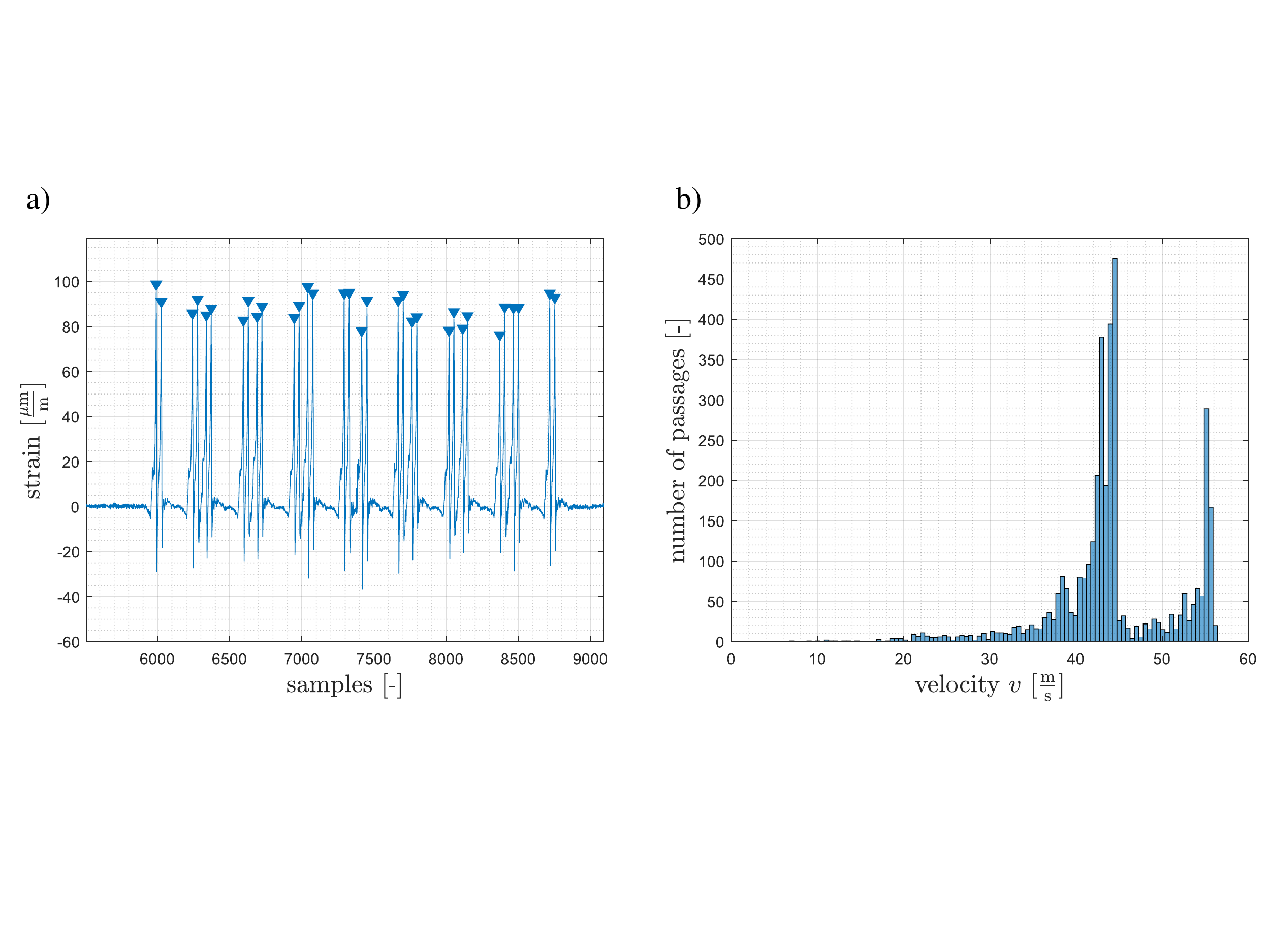}
\caption{a) Signal of wheel load measurement point with detected peaks. b) Histogram of determined mean train velocities for all 3745 passages.}
\label{F:ALMSignal_vhist}
\end{figure*}

The peaks of the wheel load measurement signals are automatically identified (fig.~\ref{F:ALMSignal_vhist}a). All passages where the two wheel load measuring points have not detected the same number of peaks are discarded. This leads to 3745 usable out of a total of 3787 recorded passages, i.e. about 98.9~\%. Using the temporal differences of the peaks at the two measuring points and the known distance between the wheel load measurement points of 14.40~m, the mean velocity can be determined for each axle. The trains reach a maximum velocity of about 57~$m/s$ (fig.~\ref{F:ALMSignal_vhist}b)

\begin{figure}[H]
\centering\includegraphics[trim = 40mm 40mm 40mm 10mm,clip,width=\linewidth]{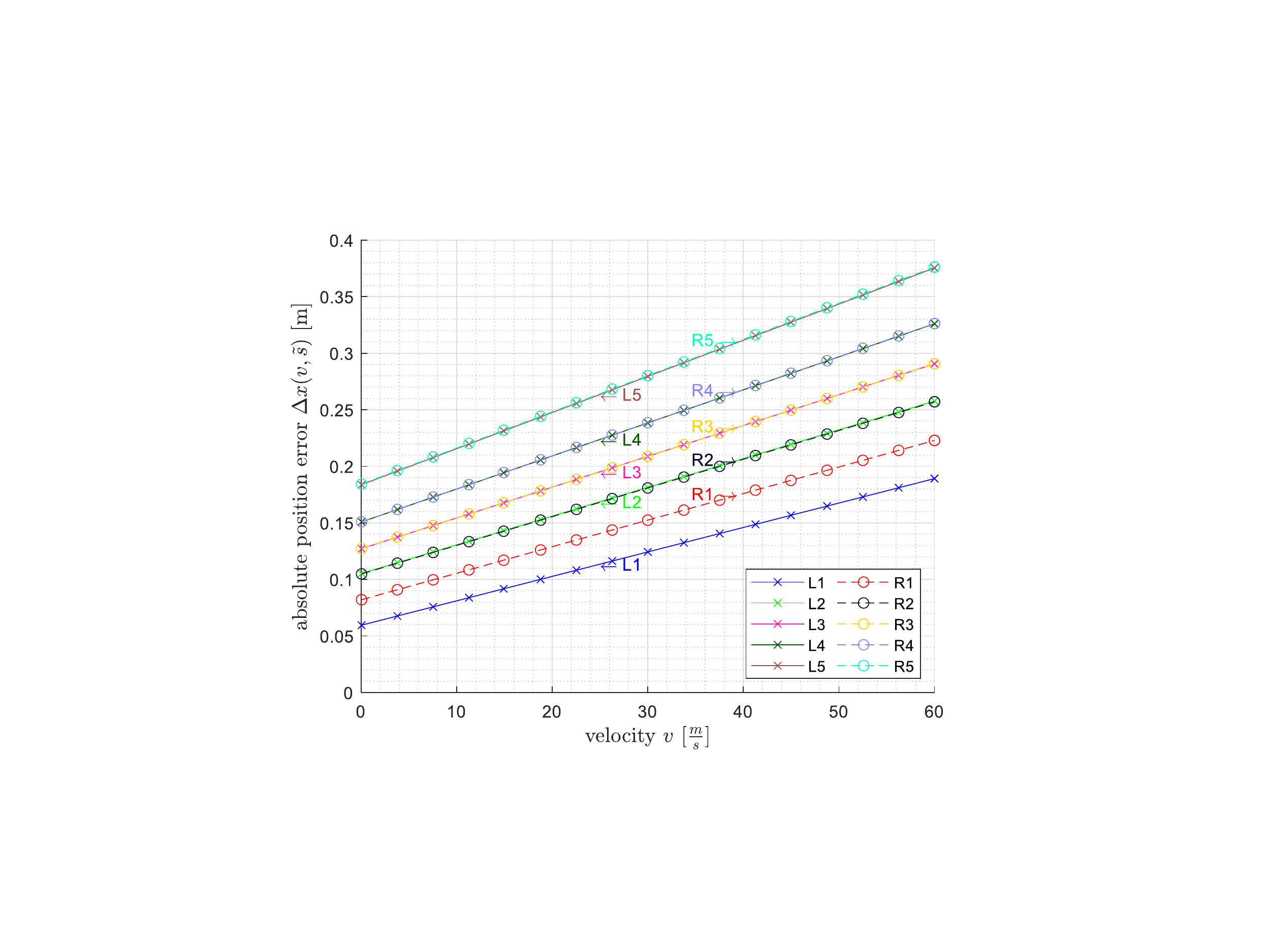}
\caption{Absolute spatial error for the calculated times at which an axle is at the same $x$-ordinate as the sensor for each of the ten sensors as a function of the train velocity}
\label{F:deltaSvsv}
\end{figure}

In the next step, using the known distances from the first wheel load measurement point to each of the ten accelerometers and the mean velocity of each axle, the time at which the axle is at the same $x$-ordinate as the respective sensor can be calculated. Since the two strain gauges of one wheel load measuring point have a distance of 20~cm between them, the uncertainty with respect to the distance between the two wheel load measuring points $s_{\mathrm{WLM}}=14.40~\mathrm{m}$ is assumed to be $\Delta s_{\mathrm{WLM}}=0.2~\mathrm{m}$. This propagates through the velocity determination. Together with an uncertainty in time of $\Delta t=\frac{1}{f_{\mathrm{s}}}=\frac{1}{600}~\mathrm{s}$, the absolute spatial error $\Delta x$ results from the linear error propagation for each sensor (fig.~\ref{F:deltaSvsv}) as follows:
\begin{equation}\label{eq:LinErrProp}
\Delta x\left(v,\tilde{s}\right)=v\Delta t+\tilde{s}\left(\lvert\frac{v}{s_{\mathrm{WLM}}}\rvert\Delta t+\lvert\frac{1}{s_{\mathrm{WLM}}}\rvert\Delta s_{\mathrm{WLM}}\right)
\end{equation}

This shows that the absolute position error is increased with increasing velocity and with an increasing distance of the sensor with respect to the first wheel load measurement point (G1/G2).

The acceleration signals were combined into one data matrix $\textbf{A}_{\mathrm{L}}^{36,000 \times 5}$ for the sensors L1 to L5 and $\textbf{A}_{\mathrm{R}}^{36,000 \times 5}$ for the sensors R1 to R5 (fig.\ref{F:BridgeSensorsetup}~b) for each passage, without any further signal processing steps. Additionally two data matrices $\textbf{L}_{\mathrm{L}}^{n_\mathrm{a} \times 5}$ and $\textbf{L}_{\mathrm{R}}^{n_\mathrm{a} \times 5}$ ($n_\mathrm{a}$: number of axles) containing the calculated indices at which an axle is at the respective sensor, are created. 

The complete data set as well as the processing code is available online \citep{Data_VADer}.

\subsection{Data Transformation}
Transforming a signal into the frequency-time domain enables the localisation of frequency content in time \citep{brunton_kutz_2019}. In our case, low-frequency effects such as the bridge natural vibration are separated from high-frequency effects such as measurement noise in the frequency domain, while the time domain is preserved. Therefore, the model can learn frequency-specific information, which should lead to faster training and more reliable results.

The most common choices for a frequency-time domain transformation are Short Time Fourier Transformation (STFT) and Continuous Wavelet Transformation (CWT). The multi-resolution approach of the CWT is particularly useful for complex signals, since it adapts the window size to the frequency \citep{Mallat1989ATF}. The STFT has a fixed resolution, which means that there is always a trade-off between a good time resolution and a good frequency resolution, depending on the window size \citep{brunton_kutz_2019}. As a result, we have chosen the CWT because it is more suitable for the analysis of acoustic and visual signals than the windowed Fourier transform \citep{DaubechiesCWT}. The CWT has also been shown in previous work to be an effective tool for axle detection \citep{Chatterjee2006, Kalhori2017, Yu2017, Zhao2020, Zhu2021}.

\begin{figure*}
    \begin{subfigure}{\textwidth}
            \includegraphics[trim = 0 13 0 0, clip, width=\textwidth] {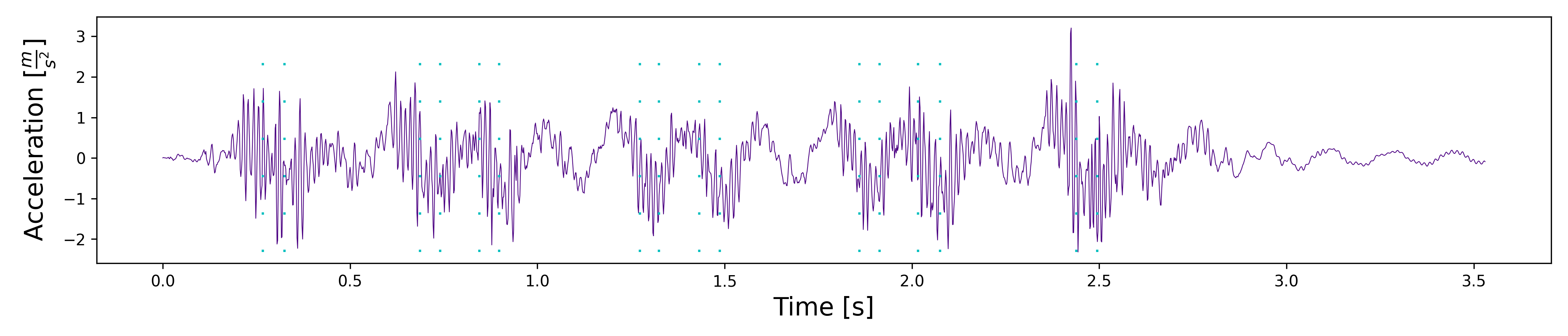}
        \caption{Acceleration signal of a single train passage}
        \label{fig:signal}
    \end{subfigure}
    \begin{subfigure}{0.495\textwidth}
            \includegraphics[trim = 0 13 0 0, clip, width=\linewidth]{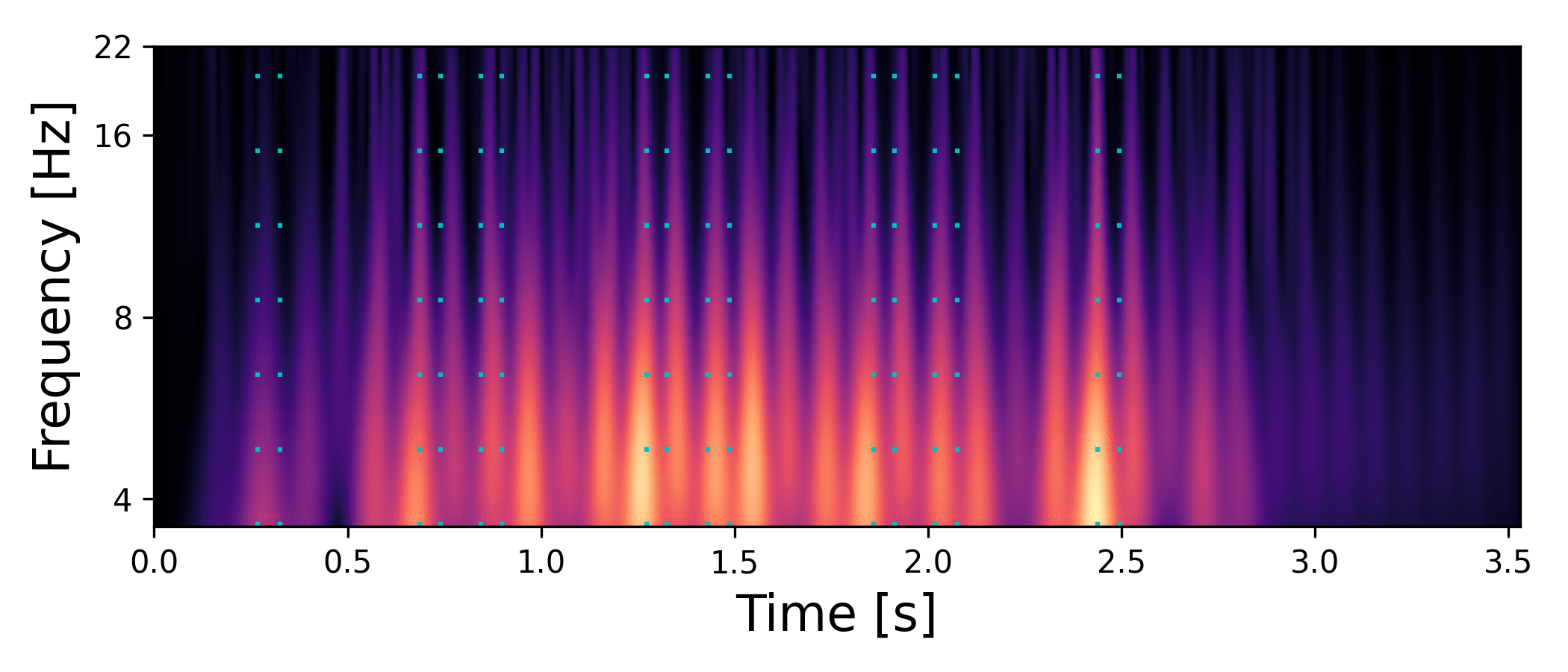}
        \caption{Complex Gaussian CWT in frequency range of bridge}
        \label{fig:cgau1}
    \end{subfigure}
    \begin{subfigure}{0.495\textwidth}
            \includegraphics[trim = 0 13 0 0, clip, width=\linewidth]{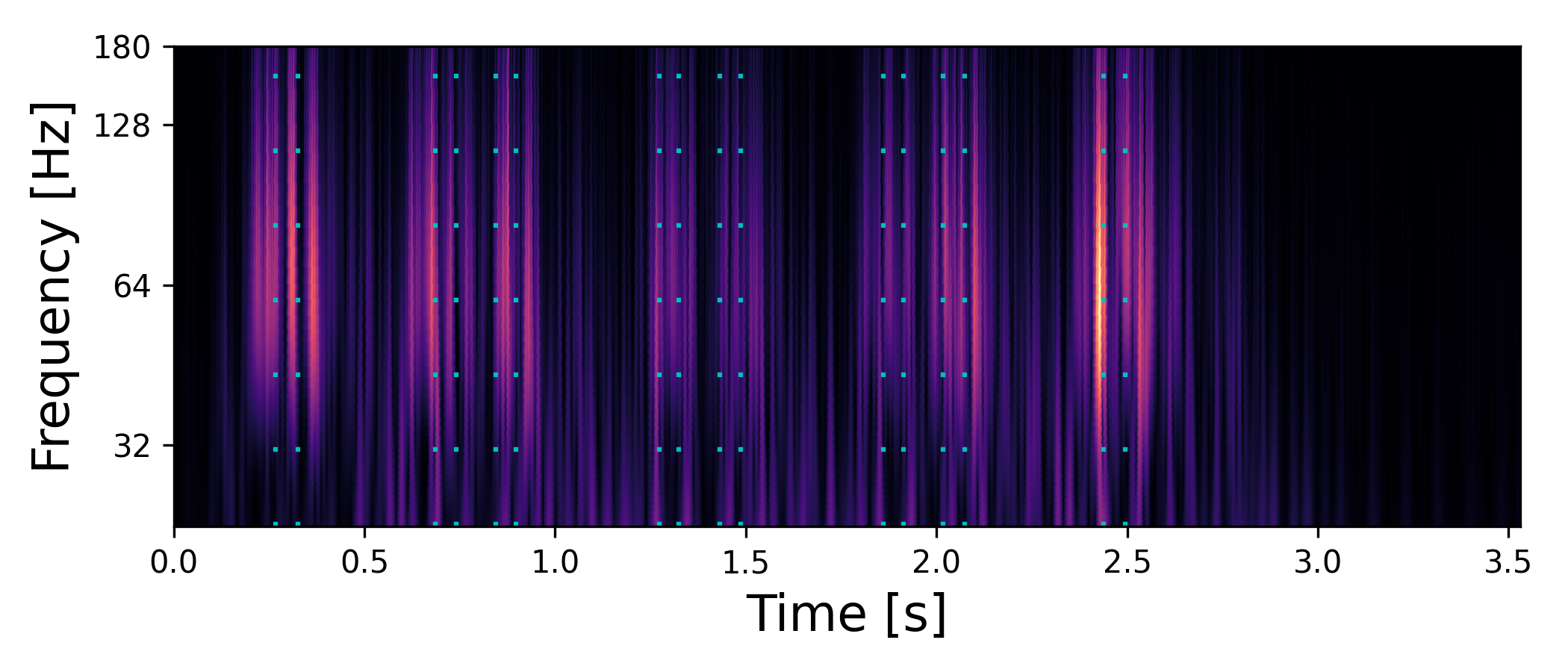}
        \caption{Complex Gaussian CWT in frequency range of axles}
        \label{fig:cgau2}
    \end{subfigure}
    \begin{subfigure}{0.495\textwidth}
            \includegraphics[trim = 0 13 0 0, clip, width=\linewidth]{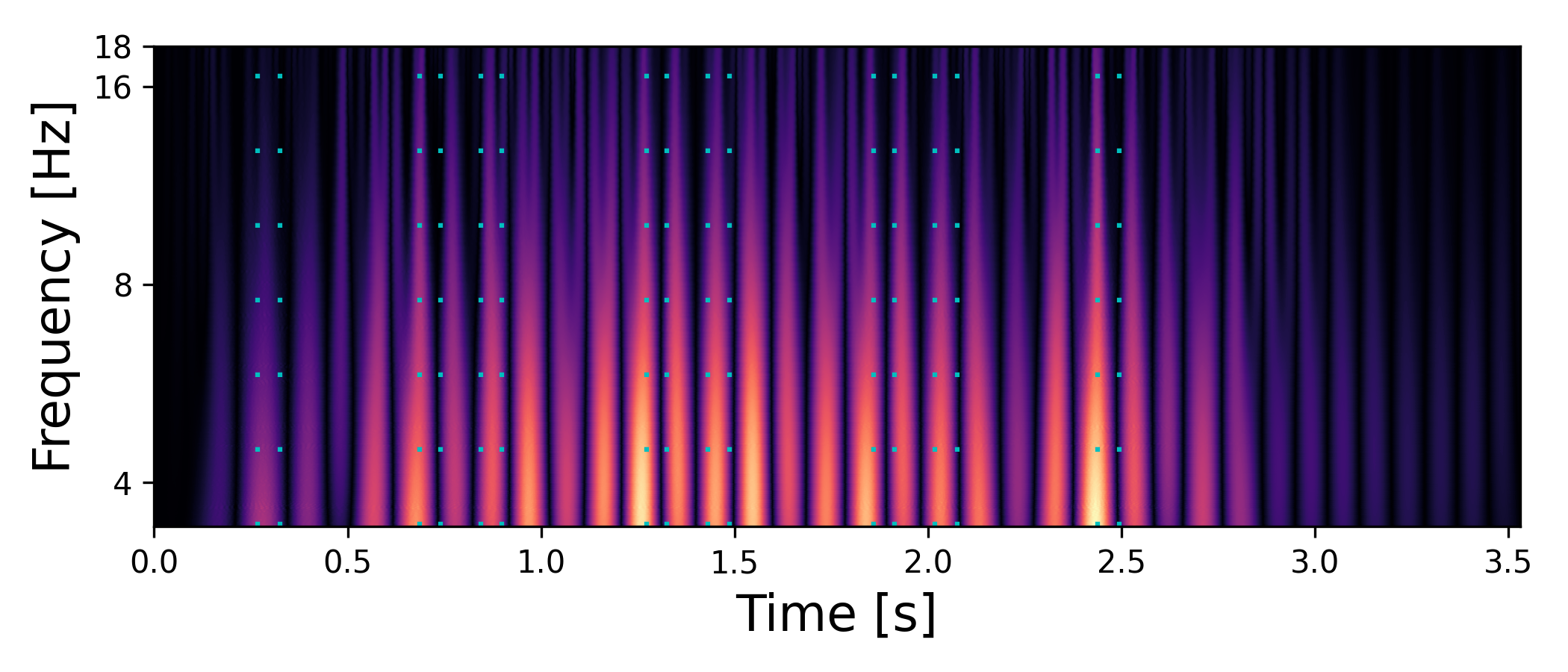}
        \caption{Gaussian CWT in frequency range of bridge}
        \label{fig:gaus1}
    \end{subfigure}
    \begin{subfigure}{0.495\textwidth}
            \includegraphics[trim = 0 13 0 0, clip, width=\linewidth]{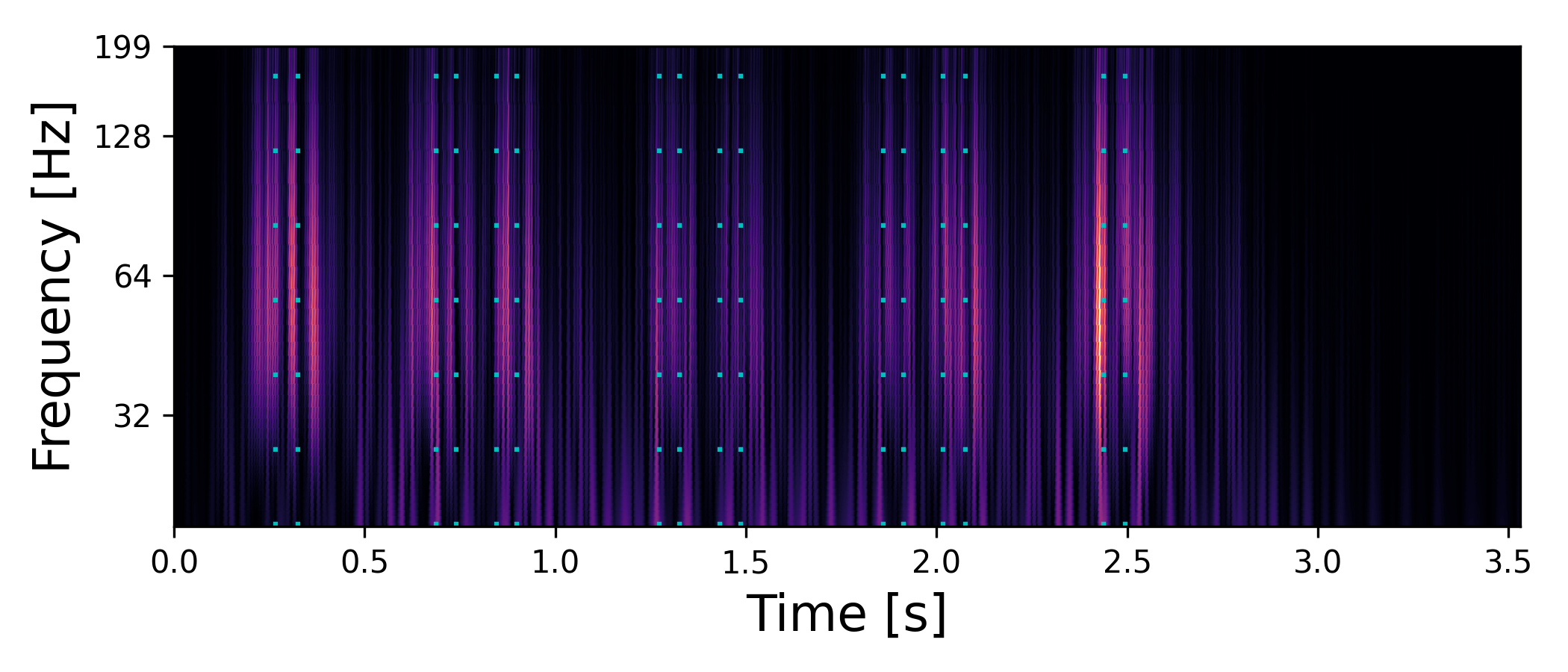}
        \caption{Gaussian CWT in frequency range of axles}
        \label{fig:gaus2}
    \end{subfigure}
    \begin{subfigure}{0.495\textwidth}
            \includegraphics[trim = 0 13 0 0, clip, width=\linewidth]{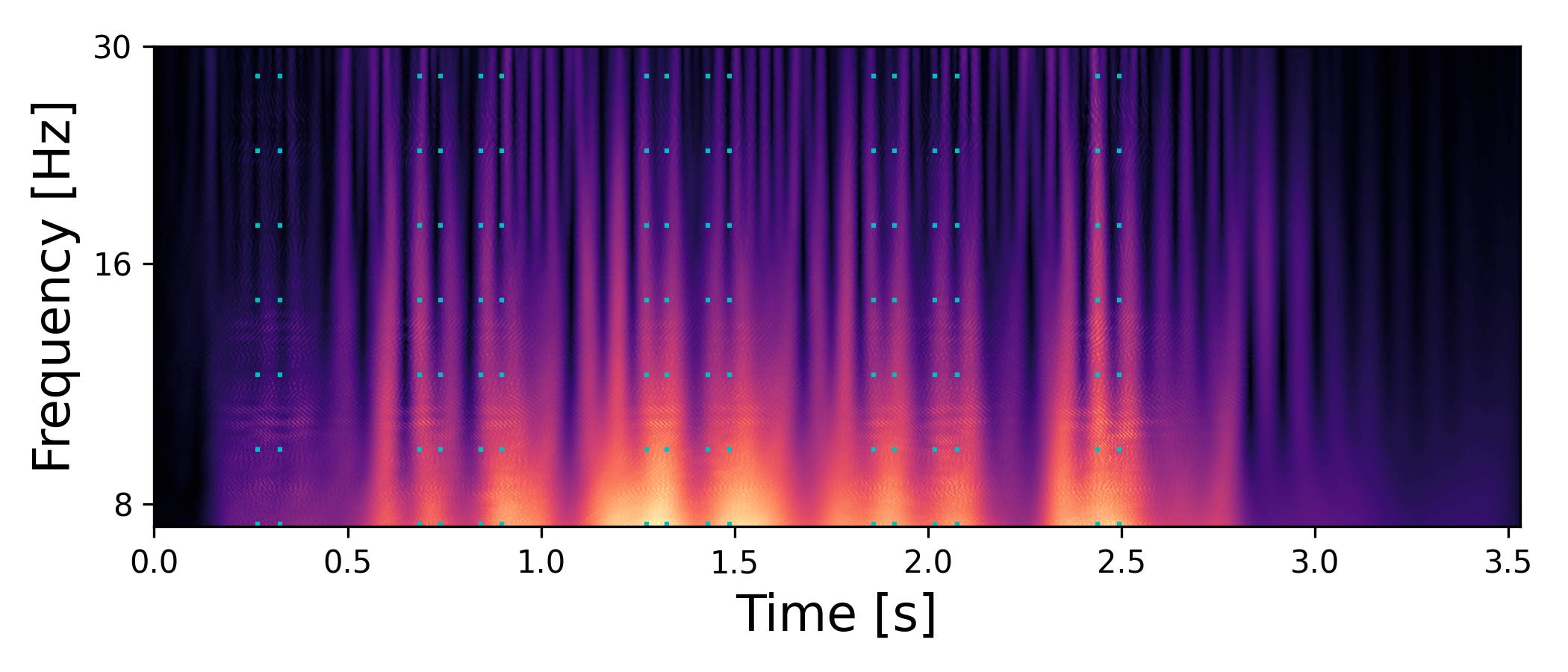}
        \caption{Frequency B-Spline CWT in frequency range of bridge}
        \label{fig:fbsp1}
    \end{subfigure}
    \begin{subfigure}{0.495\textwidth}
            \includegraphics[trim = 0 13 0 0, clip, width=\linewidth]{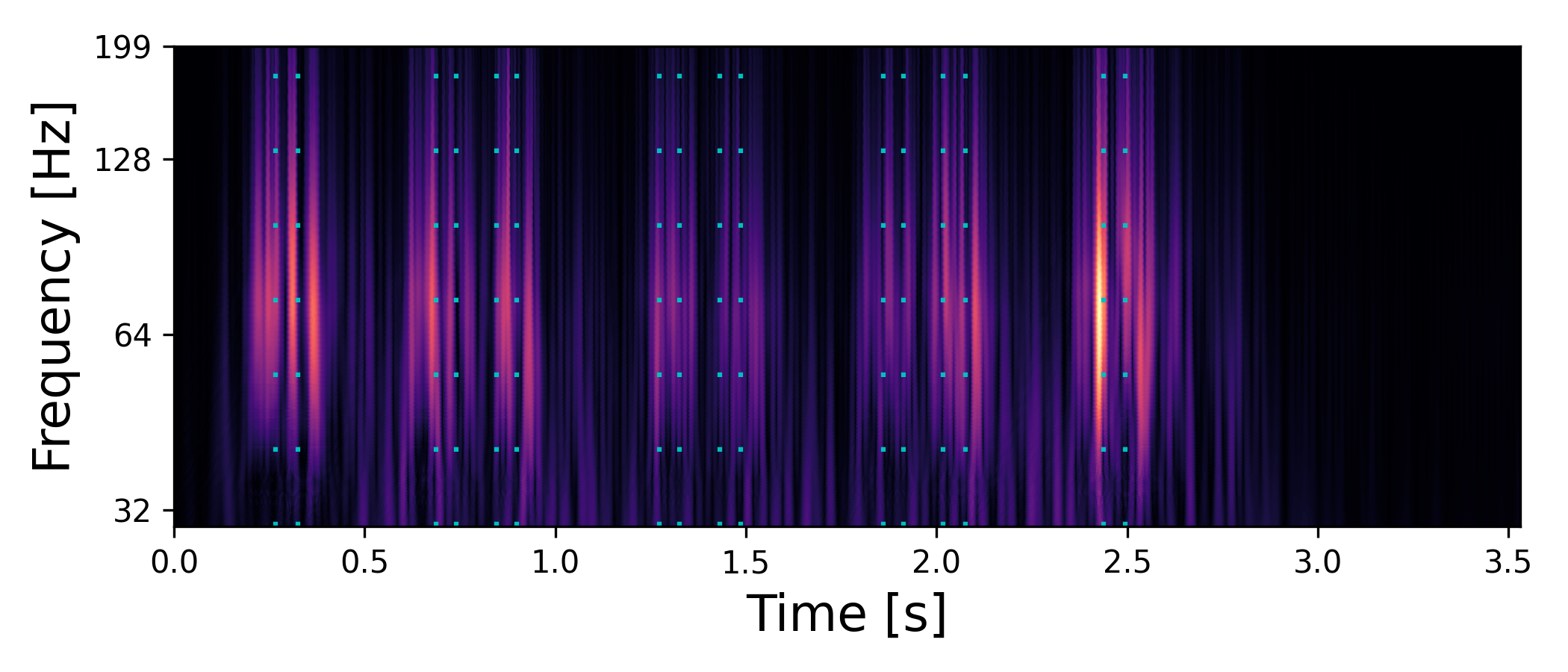}
        \caption{Frequency B-Spline CWT in frequency range of axles}
        \label{fig:fbsp2}
    \end{subfigure}
    \centering
    \caption{Continuous wavelet transformations (CWTs) for a single train passage and a single sensor. The point in time of a load transition is represented by a dashed line in cyan. The transformations were each independently normalised from 0 to 1, visualised with black for 0 and yellow for 1.}
    \label{fig:transformation}
\end{figure*}

From the signals, a section of 150 samples before the first axle to 500 samples after the last axle was further processed and transformed with the PyWavelets package \citep{Lee2019} with empirically determined settings (tab.~\ref{tab:cwtsettings}). To determine the transformation settings, the CWT were visualised and analysed for correlations between the axle positions (cyan dotted line) and the power of the transformed signal (fig.~\ref{fig:transformation}). As a result, we can find that in the range of the bridge's natural frequency of about 6.9~Hz for the first bending mode (fig.~\ref{fig:transformation} left column), the influence of the bridge on the vibration is mainly visible while a correlation between the train axles (dashed cyan lines) and the signal seem not to be present. In the higher frequency range, a correlation becomes clearer, indicating that the influence of the axles are mainly located in the~64 Hz range (fig.~\ref{fig:transformation} right column).

We assume, however, that it is nevertheless advantageous for the model to receive both pieces of information (influence of the bridge and of the axles) in order to be able to distinguish them better. 

As a result, all 6 transformations were used in combination (fig.~\ref{fig:cgau1}-\ref{fig:fbsp2}). To create the final model inputs, each signal (per passage and per sensor) was transformed according to our 6 settings, afterwardsthe transformations were normalised independently and stacked into a three-dimensional array $\textbf{T}^{n_{\mathrm{s}} \times n_{\mathrm{f}} \times n_{\mathrm{t}} }$ ($n_\mathrm{s}$: number of samples, $n_\mathrm{f}$: number of frequencies/scales and $n_\mathrm{t}$: number of transformations). 

\begin{table}
\begin{center}
\begin{tabular}{ l|c|c|c } 
    Wavelet & Figure & \makecell{Lower \\ Scale \\ Limit} &  \makecell{Upper \\ Scale \\ Limit} \\
    \hline \hline
    \multirow{2}{10em}{First Order Complex Gaussian Derivative} & \ref{fig:cgau1} & 1 & 8 \\ 
    & \ref{fig:cgau2} & 8 & 50 \\ 
    \hline
    \multirow{2}{10em}{First Order Gaussian Derivative} & \ref{fig:gaus1} & 0.6 & 6.5 \\ 
    & \ref{fig:gaus2} & 6.5 & 35 \\ 
    \hline
    \multirow{2}{10em}{Default Frequency B-Spline \citep{Lee2019}} & \ref{fig:fbsp1} & 1.5 & 10 \\ 
    & \ref{fig:fbsp2} & 10 & 40 \\ 
    
\end{tabular}
\end{center}
\caption{Continuous wavelet transformation settings}
\label{tab:cwtsettings}
\end{table}

\subsection{Model Definition}
For a Virtual Axle Detector (VAD) to be efficient, a model with a flexible input length (in the time domain) is essential to account for the large differences in velocities and train lengths. Therefore, we have developed a Fully Convolutional Network (FCN) \citep{Long2015FullyCN}, which only uses input size independent layers like convolution, pooling or batch normalization. Our model has been developed to output only a single value between 0 and 1 for the same number of samples as the input. These output values represent the model's certainty for an axle at $x$-ordinate of the respective sensor. 

Our developed VAD model is based on the U-Net architecture originally proposed by \citet{ronneberger2015unet}, which was developed for semantic segmentation tasks. Here, the goal is to classify each pixel of the input image individually to preserve the resolution from the input. For the U-Net, the resolution of the input is halved 4 times (via max pooling) in the encoder path and then doubled again 4 times (via transposed convolution) in the decoder path. In addition, the intermediate results before each pooling layer are appended to the intermediate results after the transposed convolution layer with the same resolution and then processed together.

\begin{figure*}
    \centering
    \includegraphics[width=\linewidth, trim= 46 68 8 0,clip]{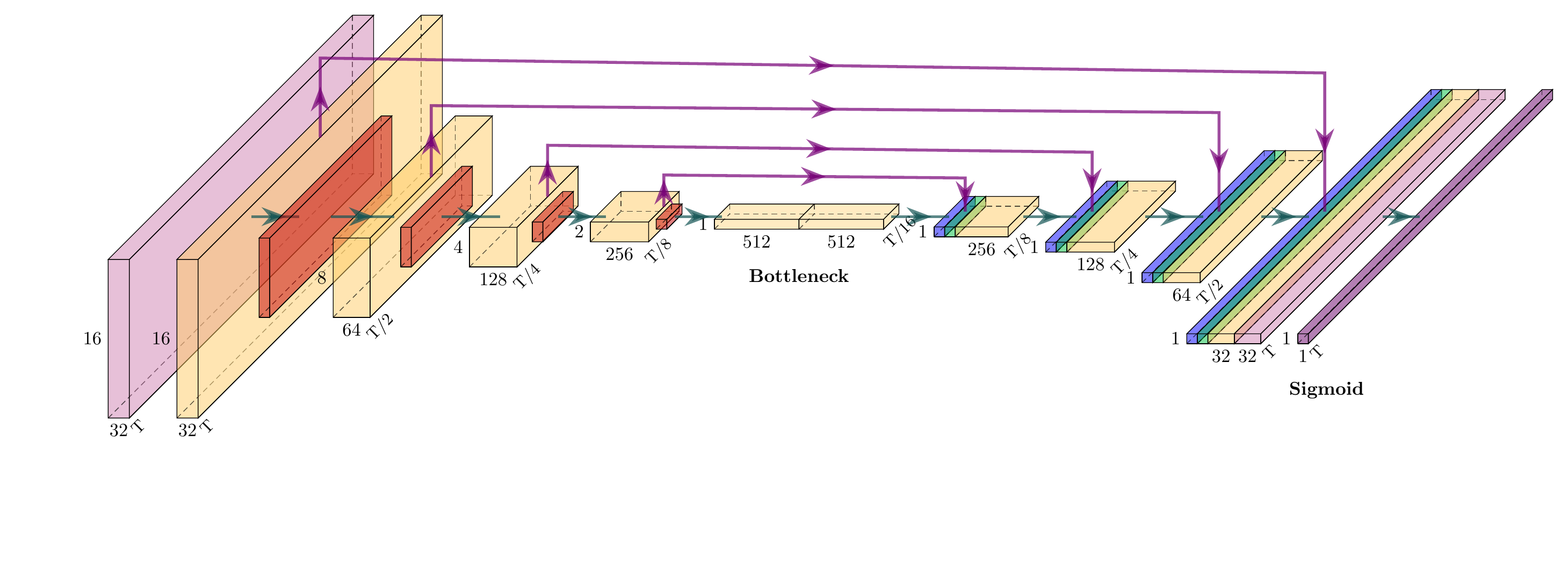}
    \caption{Definition of the Virtual Axle Detection model (VAD) with colored boxes corresponding to the following layers: convolution block (light purple), residual block (yellow), max pooling (red), concatenate (green), transposed convolution (blue), classification block (purple) and reshaping skip connection (purple arrow). Dimension of the feature maps at the corresponding boxes with T samples at the bottom right, feature maps at the bottom and frequencies at the left.}
    \label{fig:VAD}
\end{figure*}

In our case, not each pixel but each sample is to be classified, thus reducing the resolution in the frequency domain to 1. We achieve this by increasing the resolution in the decoder path only in the time dimension (fig.~\ref{fig:VAD}) by using a transposed convolution layer with a kernel size of \(3 \times 1\). Before the intermediate results from the encoder path can be appended to the intermediate results from the decoder path, its resolution and number of feature maps are adapted. Each purple arrow in fig.~\ref{fig:VAD} consists of a reshape layer to reduce the frequency domain to 1 value, and a convolution layer with \(1 \times 1\) kernel size to adapt the number of feature maps. 

The convolution blocks (CBs) consist of a batch norm layer and a convolution layer with Rectified Linear Unit (ReLU) activation \citep{geron}. The CBs in fig.~\ref{fig:VAD} have a \(3 \times 3\) kernel size. The residual blocks originally proposed by \citet{he2015deep} were implemented consisting of 3 CBs in the filtering path and 1 CB in the skip connection. Here, the second CB in the filtering path has a \(3 \times 3\) kernel size, while the other CBs have a \(1 \times 1\) kernel size. The results of the filter path and the skip connection are added element wise before they are further processed. Our model has 4 pooling steps as the U-Net \citep{ronneberger2015unet}. We can therefore input transformed signals of any length (in the time domain) as long as they are divisible by 16, since the resolution must remain an integer after being halved 4 times. For lengths that are not multiples of 16, the signal is padded with zeros and thus extended by a maximum of 15 samples. 

The last layer is a convolution layer with a single kernel of size \(3 \times 3\) with sigmoid activation. Therefore, the resulting outputs can be interpreted as independent pseudo probabilities \(p\), which indicate the predicted likeliness for a certain class per sample. The resulting model has an input size of arbitrary number of samples (padded to a multiple of 16), arbitrary number of signal transformations and 16 frequencies, evenly spaced from minimum to maximum scale. The TensorFlow library \citep{tensorflow2015-whitepaper} is used for implementation of the model and PlotNeuralNet \citep{haris_iqbal_2018_2526396} was used for visualising it.

\subsection{Loss Function}
We have defined the localisation task as a supervised classification problem instead of a regression problem in order to minimise complexity and maximise comprehensibility. We have labeled each sample with one of the following classes: Axle at the same $x$-ordinate as the sensor (1) or not (0).


A common loss function for a binary classification task is Cross Entropy (CE), but for imbalanced datasets Focal Loss (FL) has been shown to be more effective \citep{Lin2017FocalLF}. In our case, the total number of axles of a train is almost negligible compared to the total amount of samples of a passage. So if the model predicts all values to be 0 (and would not locate an axle), it would already achieve an almost perfect loss for CE and would learn to ignore the axles. This brings us to the thesis that FL should be necessary to achieve good results. The FL is defined as follows \citep{Lin2017FocalLF}:



\begin{equation}\label{eq:fl}
\mathrm{FL}(p_\mathrm{t})=-1(1-p_\mathrm{t})^\gamma\log(p_\mathrm{t}),
\end{equation}

where \(p_\mathrm{t}\) is defined as following:

\begin{equation}\label{eq:pt}
p_\mathrm{t} =
    \begin{cases}
        p & \mathrm{if}~y = 1\\
        1 - p & \mathrm{otherwise.}
    \end{cases}
\end{equation}

In the above \(p \in [0,1]\) is the model's estimated probability for the class 1, \(y\) is the ground-truth class and \(\gamma\) is the focusing parameter. The equation of FL consists of \(-\log(p_\mathrm{t})\) which is equal to the CE and \((1-p_\mathrm{t})\) which is a newly introduced modulating factor weighted by the focusing parameter \(\gamma\). The larger the factor, the more significant is the effect of the modulating factor and with a \(\gamma\) of 0 FL corresponds to the CE \citep{Lin2017FocalLF}.

Due to the gamma value, the modulating factor is included exponentially in the equation. As a result, the loss becomes exponentially smaller the better the prediction. For misclassified examples, the loss is unaffected compared to CE, which makes misclassifications much more heavily weighted (a factor of 1000 and more is possible \citep{Lin2017FocalLF}).




\subsection{Evaluation Metrics}

The loss function itself does not contain information about the number of correctly detected axles. Other metrics are needed to assess the overall performance of the VAD. Accuracy as a metric is also insufficient to draw a conclusion about the models performance, due to the imbalance of our dataset. A prediction containing no axles at all would reach an accuracy of about 99\% and would therefore not contain useful information. Precision and recall are suitable metrics for imbalanced data sets \citep{geron}, but they only take into account binary results and not distance prediction and ground-truth. Due to the high sampling rate and the uncertainty of the labels described in section~\ref{S:2}, however, we want to recognise axle predictions within a few samples next to the ground-truth as correct and measure the temporal error.

\begin{figure}
    \centering    \includegraphics[width=\linewidth]{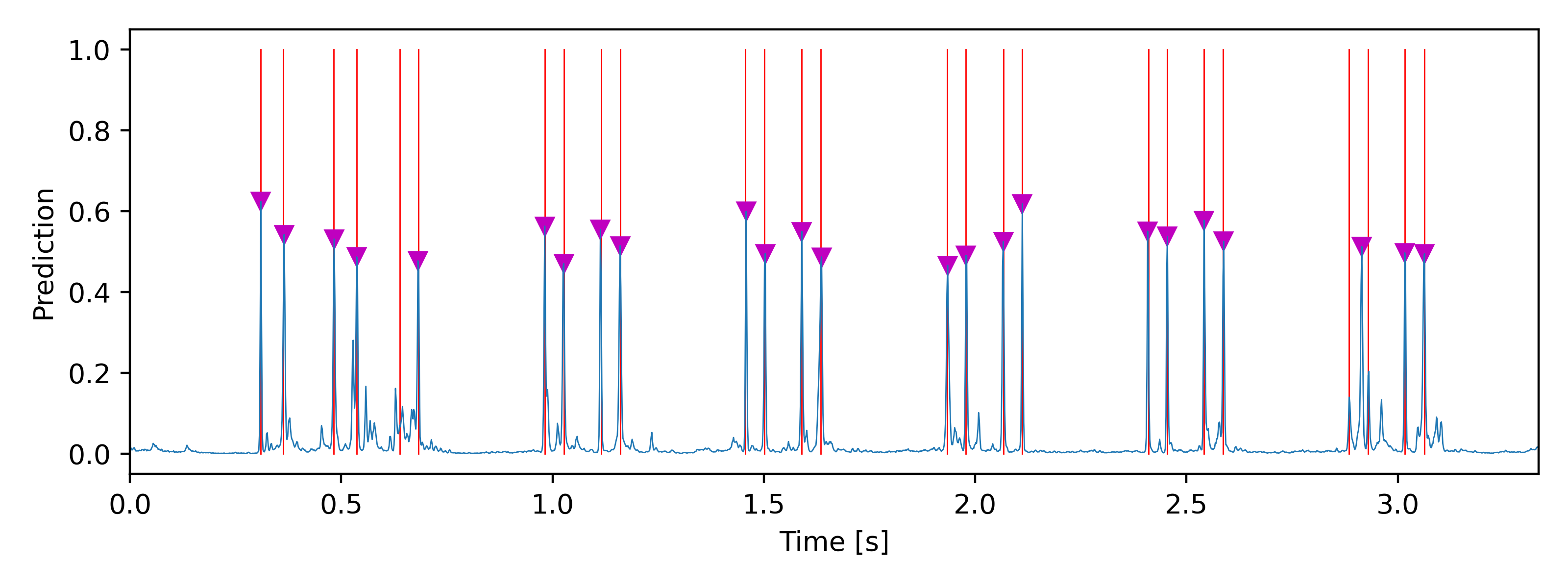}
    \caption{Exemplary output with the blue line for the model output, the red lines for the labeled axis locations and magenta triangles for the found peaks.}
    \label{fig:output}
\end{figure}

Since the model output does not always give clear results, but sometimes also a number of smaller peaks, the prediction must be processed further (figure~\ref{fig:output}). In order to ignore small values and only continue with plausible predictions, peaks are extracted in post-processing with the find-peaks function from SciPy \citep{2020SciPy-NMeth}. In order to get consistent and satisfying results, we have fine-tuned the following parameters of the function: Minimum height of the peak (0.25), minimum distance between two peaks (20~samples) and prominence of the peak compared to the surrounding points (0.15). We have calculated the minimum distance $d$ between two peaks with assumed minimum wheel distance $\Delta w_ {\mathrm{min}}=2~\mathrm{m}$ and the maximum velocity $v_{\mathrm{max}} =220~\frac{\mathrm{km}}{\mathrm{h}}$ as follows:

\begin{equation}\label{eq:dis}
d = \frac{\Delta w_{\mathrm{min}}\cdot f_\mathrm{s}}{v_{\mathrm{max}}} 
\approx \frac{2~\mathrm{m} \cdot 600~ \frac{\mathrm{samples}}{\mathrm{s}}}{61.1~ \frac{\mathrm{m}}{\mathrm{s}}}
\approx 20~\mathrm{samples}
\end{equation}


A threshold is used to ensure that only predictions within a certain temporal error compared to the ground-truth are considered correct. For example, the threshold could classify predicted axles as correct with a maximum temporal error of 30 milliseconds compared to the ground-truth. Depending on the application, its requirements may be decisive for the determination of the threshold. In general, it should be taken into account that good results cannot be expected with thresholds that are lower than the label and measurement accuracy. To avoid making too strict assumptions, we have chosen the largest reasonable threshold with 20~samples (eq.~\ref{eq:dis}) for the first evaluations. After the peaks found have been classified as correct or incorrect, they are further evaluated using the following metrics: Precision, recall and \(F_1\) score, which is the harmonic mean of precision, recall.




\subsection{Optimization of \(\gamma\)}

In order to find an optimal \(\gamma\) value for the FL, we performed a parametric study with 150 epochs per run, 150 steps per epoch and 16 samples per batch. We have split the dataset randomly with 70\% for training, 20\% validation and 10\% for testing. To ensure comparability, the same random state was used for every run. The selection criterion for \(\gamma\) is the \(F_1\) score, because a high \(F_1\)-score indicates a high value for both recall and precision.

\begin{figure}[h]
    \centering
    \includegraphics[width=\linewidth]{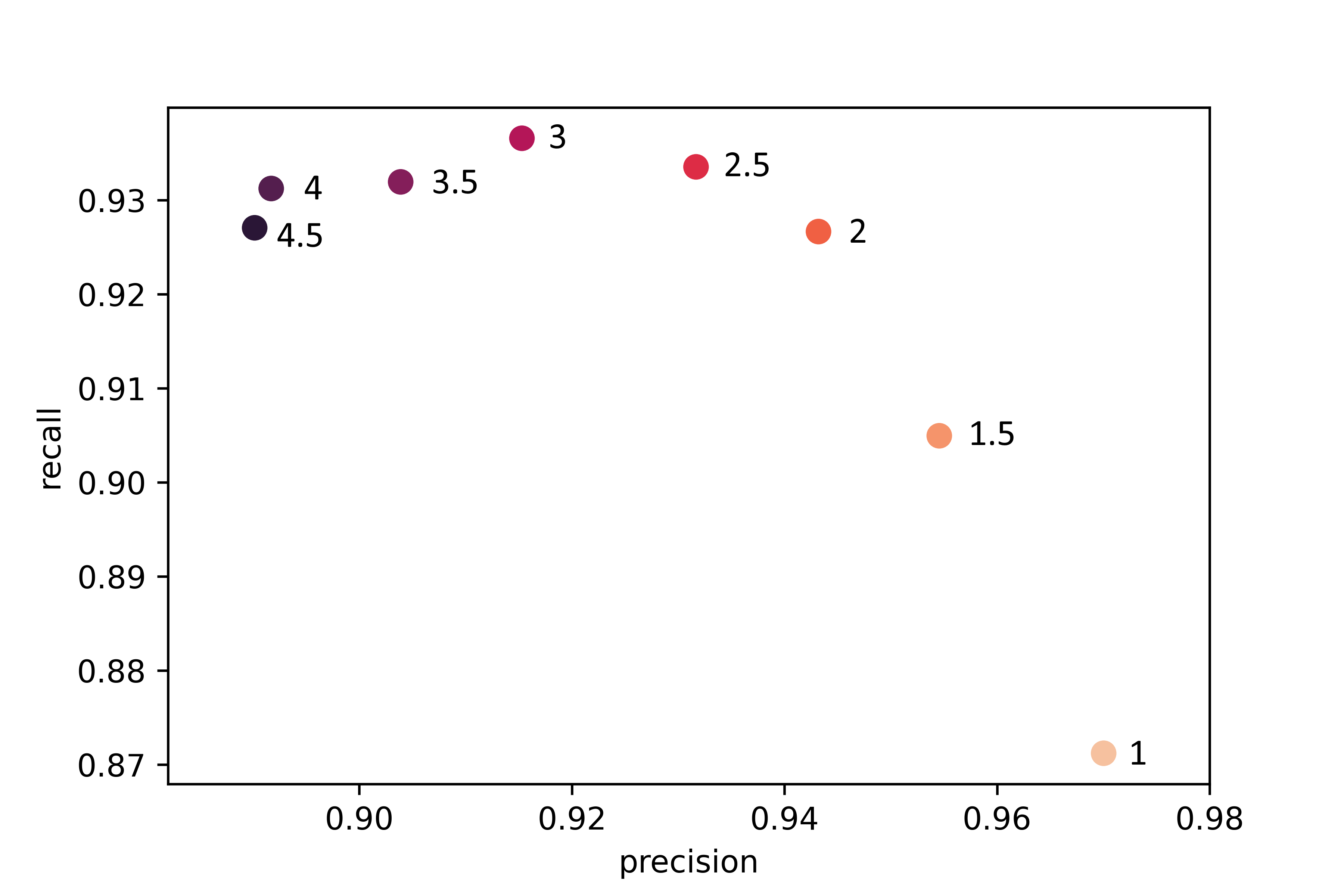}
    \caption{Relationship between \(\gamma\), precision and recall with median values of training results on validation set.}
    \label{fig:gamma}
\end{figure}

We confirmed our hypothesis that our dataset is too unbalanced for standard loss functions like Cross Entropy. The model training with small \(\gamma\) values of 0 and 0.5 ended in dead ReLUs after 8 or 9 epochs and is therefore unusable.  However, the modulation factor should also not be weighted too high to achieve the best performance. The relationship between \(\gamma\), precision and recall can be described as a trade-off between detecting too many axles and detecting too few axles (fig.~\ref{fig:gamma}). 

The \(\gamma\) values of 2, 2.5 and 3 achieved the highest \(F_1\) score on the validation set. In order to decide which \(\gamma\) value to use for the final evaluation, we have trained the model with these \(\gamma\) values in a second run for 300 epochs. In the second run, the \(\gamma\) value of 2.5 achieved the highest \(F_1\) score (tab.~\ref{tab:gamma}) and is therefore kept for testing. Since the results of the \(\gamma\) values are close to each other and the middle \(\gamma\) value performed best, we assume that the optimal value has been found.

\begin{table}[H]
\begin{center}
\begin{tabular}{c|c c c}
    \(\gamma\) & \(F_1\) & precision & recall \\
    \hline
    3 & 0.9538 & 0.9477 & 0.9620 \\
    2.5 & 0.9544 & 0.9556 & 0.9542 \\
    2 & 0.9534 & 0.9559 & 0.9522
\end{tabular}
\end{center}
\caption{The models performance on the validation set in dependence on the \(\gamma\) value of FL  with increased training length. Precision and recall values in each case from the epoch with the highest \(F_1\) value.}
\label{tab:gamma}
\end{table}

\section{Results and Discussion}
\label{S:5}
The test set consists of 375 train passages with 13,480 axles in total. There are 10 acceleration sensors for which the individual crossing times are to be determined, resulting in 134,800 times to be localised. On the test set, for a threshold of 20~samples the VAD with a \(\gamma\) value of 2.5 achieved a \(F_1\) score of 0.938, a recall of 0.946 and a precision of 0.941. Thus, 126,449 of 134,800 crossing times were localised correctly with a maximum error of 0.033~seconds. On average, the predicted axle times had a temporal error of 1.16~samples (0.002~s) compared to the ground-truth with a standard deviation of 3.06~samples (0.005~s).

\begin{figure}
    \centering
    \includegraphics[width=\linewidth]{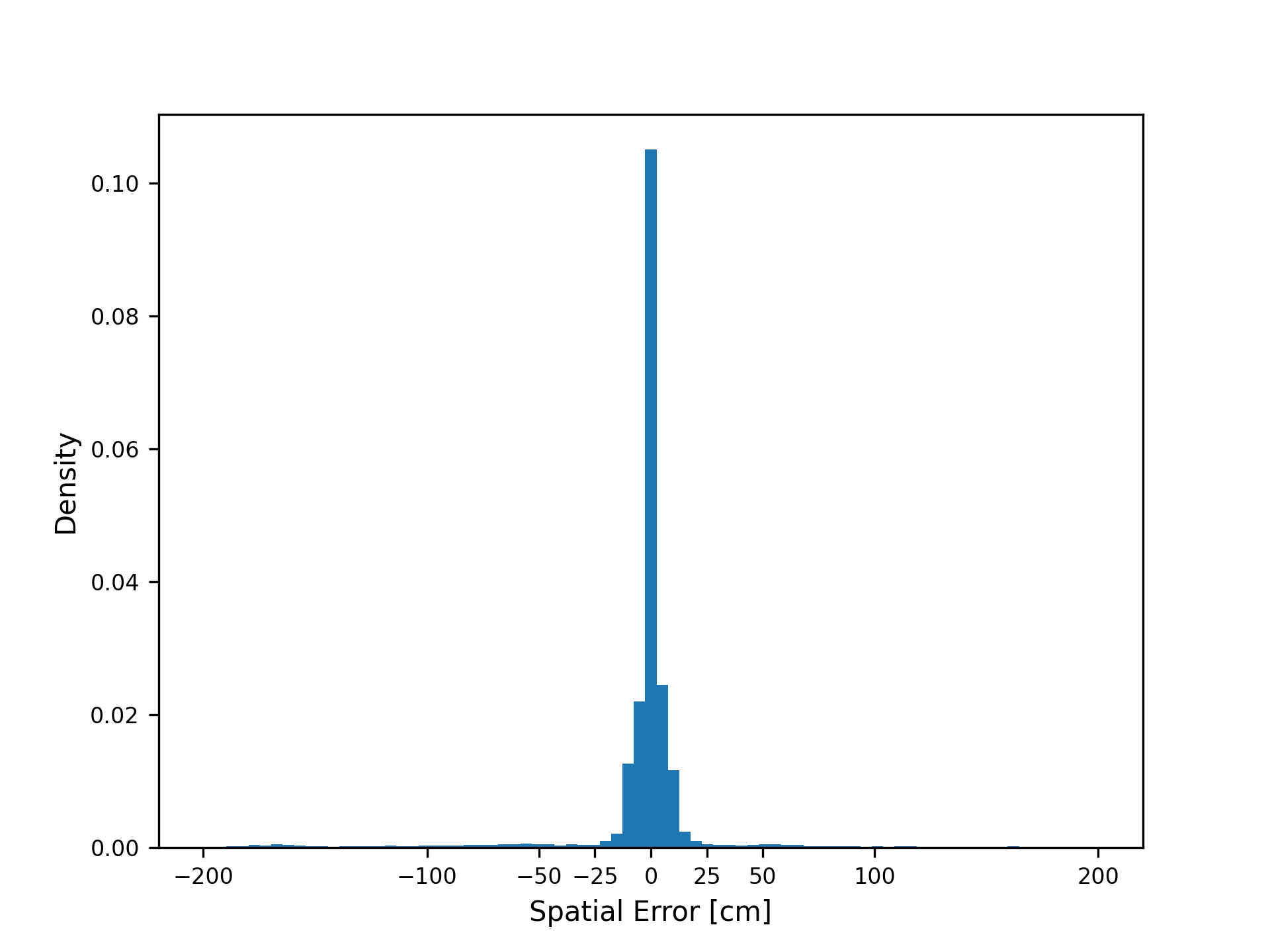}
    \caption{Differences between true and predicted axle positions for a threshold of 2m.}
    \label{fig:deviation}
\end{figure}

Based on the distances between the sensors, we are able to convert the error from samples (temporal) to meters (spatial). In order to examine the spatial error more closely, we have chosen 3 threshold values: 
\begin{itemize}
    \item 200~cm as minimum wheel distance 
    \item 37~cm as maximum labeling error  (fig.~\ref{F:deltaSvsv})
    \item 20~cm as length of wheel load measuring point
\end{itemize}
The spatial errors for a threshold of 2~m are mostly at 0~cm with an almost symmetrical distribution (fig.~\ref{fig:deviation}), thus indicating that there is no bias in the VAD. Most values are within a spatial error of 20~cm and only a few values have a higher error than 25~cm. The maximum labelling error in the velocity range of most passages (30-60~\(\mathrm{\frac{m}{s}}\)) is partially even above 25~cm (fig.~\ref{F:deltaSvsv}). 


\begin{figure}[H]
    \centering
    \includegraphics[width=\linewidth]{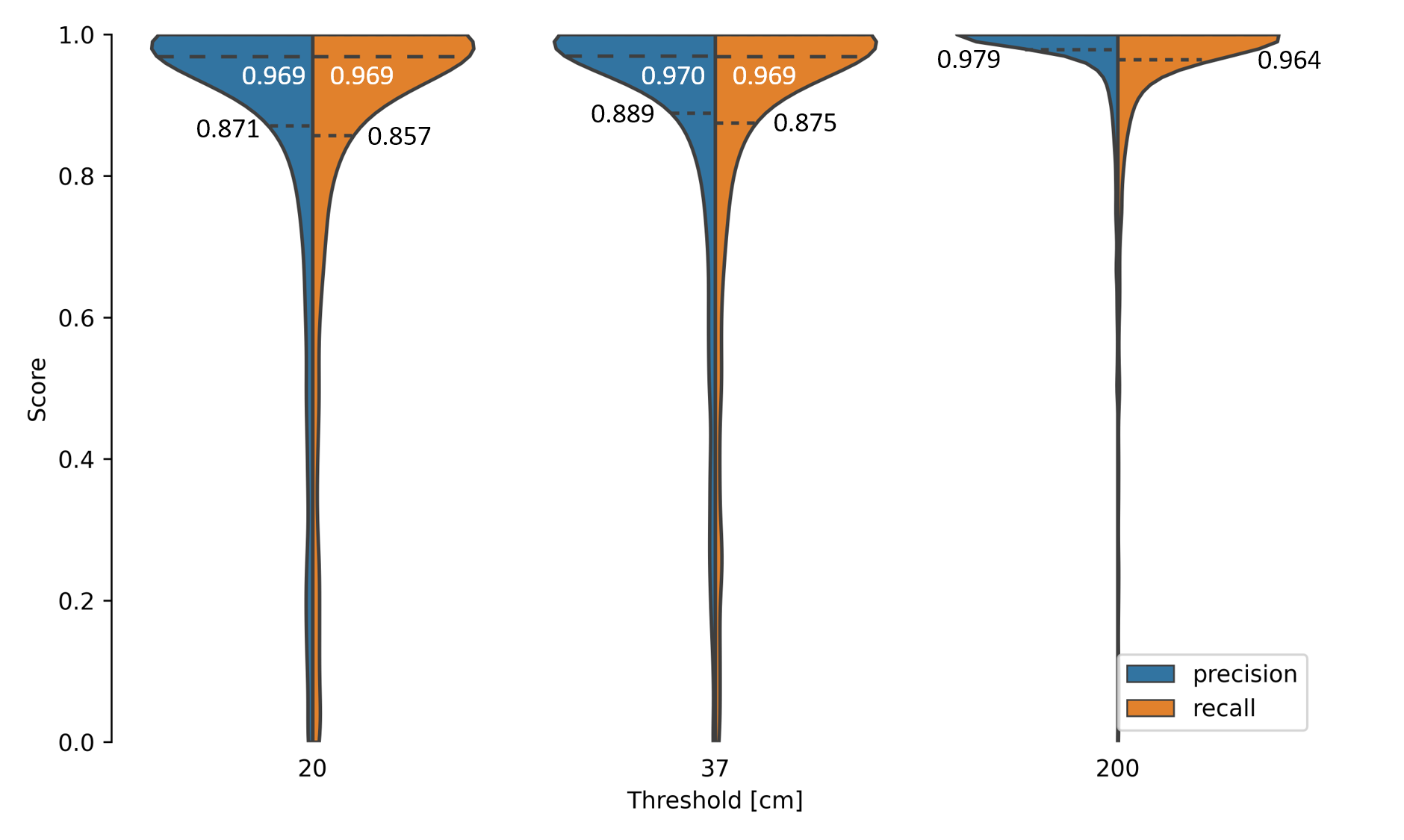}
    \caption{Precision and recall on test dataset for different thresholds. Dotted lines with black text for 25\% quantile and dashed lines with white text for median, if not at 1.0. }
    \label{fig:pr}
\end{figure}

We have calculated the precision and recall per passage and sensor for each threshold to examine the distribution of the metrics in more detail, resulting in 3750 values per threshold and metric (fig.~\ref{fig:pr}). The differences of the results with thresholds of 20~cm and 37~cm are small, as even the 25\% quantile stays above 85\% for both metrics (fig.~\ref{fig:pr}). Precision and recall for a threshold of 200~cm are much better with even the 25\% quantile above 96\%, while the mean spatial error has worsened greatly with more than double the value compared to the other thresholds (tab.~\ref{tab:mean_std}). Therefore we conclude that 37~cm is the optimal threshold value to correctly evaluate the models performance. Thus, we consider predictions with an spatial error above 37~cm as outliers. Such outliers should be possible to be sorted out in postprocessing by comparison with known train configurations. 

The evaluation of the test data took 335~seconds with an NVIDIA RTX 3090 for 375 passages and ten sensors. The model therefore needs 0.089~seconds per signal and for our entire measurement setup 0.89~seconds per passage. This allows a real-time application of the VAD and a flexible trade-off between accuracy and computing speed due to the number of sensors used.

Compared to the work of \citet{Chatterjee2006} who used FAD sensors and wavelets to detect more axles with the FAD, our model shows a comparable success rate in detecting axles. They were able to successfully evaluate 42/47 (about 89.4\%) passages. The mean absolute spatial errors are about 10.6~cm which is about three times as much as in our study. 
The achieved spatial accuracy in our study is still 1.4 times better compared to a study with FAD sensors in combination with a optimized mother wavelet and wavelet scale for the identification of axles \citep{Zhao2020}.
Taking into account that we did not use FAD in our method and the velocities are about twice as high, this is a confirmation of our hypothesis that it is advantageous not to limit the analysis to certain mother wavelets and certain scales. 
In contrast to the method of \citet{Zhu2021}, due to 
our model architecture the VAD can be applied at any point of the bridge. This allows common SHM measurement setups to be used for axle detection without the need to attach additional sensors. The accuracy of the methods is similar. It should be noted that in all cases the detection of car axles is compared with that of train axles. 

\begin{table}[H]
\begin{center}
\begin{tabular}{c|c c c c}
    threshold [cm] & mean [cm] & \(F_1\) & precision & recall \\
    \hline
    200 & 10.3 &  0.954 & 0.970 & 0.948\\
    37 & 3.9 &  0.915 & 0.926 & 0.910 \\
    20 & 3.5 &  0.897 & 0.905 & 0.892
\end{tabular}
\end{center}
\caption{Influence of the threshold on mean spatial error, \(F_1\), precision and recall.}
\label{tab:mean_std}
\end{table}

\section{Conclusion}
\label{S:6}
We were able to show that using our proposed method, no additional FADs or strain gauges are required on the main girders to realise a NOR-BWIM system. Instead, our method allows accelerometers at any point of the structure to be used as VADs.

We have shown that FCNs are able to detect axles only using acceleration measurements within a spatial accuracy of 37~cm with an precision of 93\% and recall of 91\%, thereby the mean of the absolute values of the spatial errors  compared to the ground-truth is about 3.9~cm. The results show that the method is able to detect the axles with similar spatial error as the data used for labeling. 


Even if the results show a higher accuracy compared to other studies with a different methodology, we assume, that the accuracy for the determination of the vehicle configuration and velocity could be increased by the joint evaluation of several sensors, an increased model complexity, improved signal transformation or different measured variables such as strain and displacement. Enabling the method for other measured variables would also increase the amount of use cases.

Finally, the most important issue is the generalisability of the model. Depending on whether the model needs to be re-trained for the application of the method, if so with real data or simulated data, will determine how efficiently it can be used. If retraining with real data is necessary, we suggest to determine the axle position during passages with the help of vehicles with known axle configuration and a DGPS.

\section*{Author Contributions}
\label{S:8}
\textbf{Steven Robert Lorenzen:} Conceptualization, experimental realisation and data curation, funding acquisition and writing - original draft. \textbf{Henrik Riedel:} Machine learning - model architecture \& task definition, data transformation, software, investigation, visualization and writing - original draft. \textbf{Maximilian Michael Rupp:} Formal analysis, machine learning - task definition and writing - original draft. \textbf{Leon Schmeiser:} Machine learning - task definition, writing - review \& editing. \textbf{Hagen Berthold:} Experimental realisation and data curation, funding acquisition and writing - review \& editing. \textbf{Andrei Firus:} Funding acquisition and writing - review \& editing. \textbf{Jens Schneider:} Supervision, resources and writing - review \& editing.


\section*{Acknowledgments}
\label{S:9}
The research project ZEKISS (www.zekiss.de) is carried out in collaboration with the German railway company DB Netz AG, the Wölfel Engineering GmbH and the GMG Ingenieurgesellschaft mbH. It is funded by the mFund (mFund, 2020) promoted by the The Federal Ministry of Transport and Digital Infrastructure.

The research project DEEB-INFRA (www.deeb-infra.de) is carried out in collaboration with the the sub company DB Campus from the Deutschen Bahn AG, the AIT GmbH, the Revotec zt GmbH and the iSEA Tec GmbH. It is funded by the mFund (mFund, 2020) promoted by the The Federal Ministry of Transport and Digital Infrastructure.

\begin{figure}[H]
\includegraphics[trim =55 190 120 190, clip, width=\linewidth]{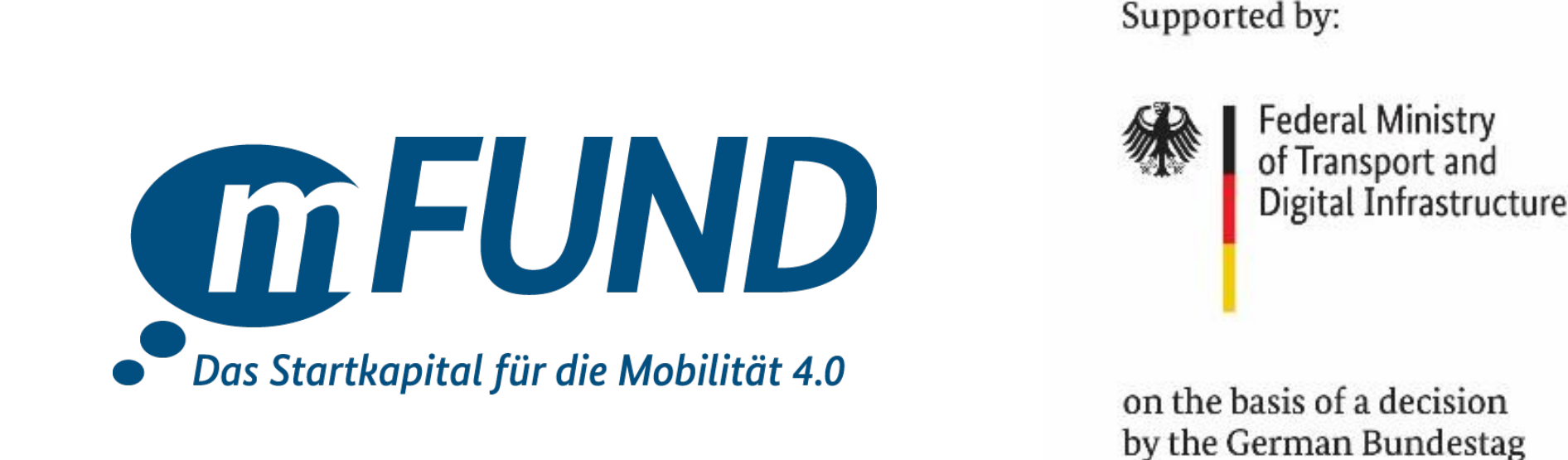}
\end{figure}

\section*{Data and Source Code}
The data \citep{Data_VADer} as well as the source code \citep{Github_VADer} used in this paper is published and contains:
\begin{enumerate}
    \item All measurement data
    \item Matlab code to label data and save as text files
    \item Python code for transformation, training, evaluation and plotting.
\end{enumerate}

\bibliographystyle{elsarticle-harv}
\bibliography{References}

\end{document}